\documentclass[letterpaper, 10 pt, conference]{ieeeconf}  

\IEEEoverridecommandlockouts                              

\usepackage[usenames,dvipsnames,table]{xcolor}

\usepackage{amsmath, amssymb, amsfonts}
\usepackage{pifont}  
\newcommand{\cmark}{\ding{51}}
\newcommand{\xmark}{\ding{55}}
\usepackage{graphicx}
\usepackage[colorlinks, urlcolor=cyan]{hyperref}
\usepackage{booktabs}
\usepackage{siunitx}
\usepackage{adjustbox}
\usepackage{tabularx}
\usepackage[T1]{fontenc}

\usepackage{subcaption}
\usepackage[nolist,nohyperlinks]{acronym}
\begin{acronym}
  \acro{AUV}{Autonomous Underwater Vehicle}
  \acro{UUV}{Unmanned Underwater Vehicle}
  \acro{CI}{continuous integration}
  \acro{DVL}{Doppler Velocity Logger}
  \acro{FTRT}{faster-than-real-time}
  \acro{LRAUV}{Long-Range Autonomous Underwater Vehicle}
  \acro{RTF}{real-time factor}
  \acro{SV}{sampling vehicle}
  \acro{RV}{relief vehicle}
  \acro{DAT}{directional acoustic transponder}
\end{acronym}

\begin{document}

\setlength{\abovedisplayskip}{4pt}
\setlength{\belowdisplayskip}{4pt}

\title{\LARGE \bf
From Concept to Field Tests: Accelerated Development of Multi-AUV Missions Using a High-Fidelity Faster-than-Real-Time Simulator
}

\author{%
    Timothy R. Player$^{*}$,
    Arjo Chakravarty$^{\dagger \star}$,
    Mabel M. Zhang$^{\dagger}$,
    Ben Yair Raanan$^{\ddagger}$, \\
    Brian Kieft$^{\ddagger}$,
    Yanwu Zhang$^{\ddagger}$,
    and Brett Hobson$^{\ddagger}$
  \thanks{$^{*}$ This author is with \textit{Oregon State University}, Corvallis, Oregon, USA. This work was conducted as a summer intern at \textit{Monterey Bay Aquarium Research Institute}.
    {\tt\small playert@lifetime.oregonstate.edu}.
  }%
  \thanks{$^{\dagger}$ These authors are with \textit{Open Robotics}, Mountain View, California, USA
    {\tt\small\{arjo, mabel\}@openrobotics.org}.
  }
  \thanks{$^{\star}$ This author is also with \textit{Singapore University of Technology and Design}, Singapore
    {\tt\small 1007419@mymail.sutd.edu.sg}.
  }
  \thanks{$^{\ddagger}$ These authors are with \textit{Monterey Bay Aquarium Research Institute}, Moss Landing, California, USA.
    {\tt\small\{byraanan, bkieft, yzhang, hobson\}@mbari.org}.
  }
}

\maketitle

\begin{abstract}
We designed and validated a novel simulator for efficient development of multi-robot marine missions. To accelerate development of cooperative behaviors, the simulator models the robots' operating conditions with moderately high fidelity and runs significantly faster than real time, including acoustic communications, dynamic environmental data, and high-resolution bathymetry in large worlds. The simulator's ability to exceed a \ac{RTF} of 100 has been stress-tested with a robust continuous integration suite and was used to develop a multi-robot field experiment.
\end{abstract}

\section{Introduction}
Autonomous robots are a mainstay of modern ocean exploration. Robots  collect measurements in situ at larger scales, with greater precision, and at significantly lower cost than traditional ship operations. Further, multi-robot systems in long-duration deployments can collect larger-scale data more efficiently than single \acp{AUV} \cite{zarokanellos2022frontal, zhang2021system, Hobson2012}. However, complex underwater multi-robot systems require rigorous validation in simulation and in the field to function reliably.

Developing long-duration multi-AUV missions is challenging because many failure modes could jeopardize success in week or month-long deployments. Underwater platforms must operate reliably with limited communication, constrained power, and uncertain localization. Failures can result in the loss of expensive payloads and data. The risk increases with multiple agents with more points of failure. 
Simulation plays a key role by allowing code to be tested before high-stake deployments. However, 
existing simulators are too slow or do not support multiple vehicles.

We designed a simulation stack, LRAUV Sim\footnote{\url{https://github.com/osrf/lrauv}}, for developing complex multi-AUV missions. The simulator can be extended to incorporate arbitrary propeller-driven underwater vehicles, but presently models \acp{LRAUV}, a slender AUV that is regularly deployed in the real world by two institutions. \ac{LRAUV} Sim extends the new Gazebo simulator \cite{newgazebo} and models hydrodynamics, acoustic communications, and marine sensors at what we believe to be greater speed than previous simulators, while allowing scientific data to be visualized from a user-provided scalar field. \ac{LRAUV} Sim enables a development paradigm for field robotics in which multiple action sequences can be quickly simulated to test failure cases and mission logic in complex systems.

\begin{figure}[t]
  \begin{subfigure}{.46\linewidth}
    \begin{center}
      \includegraphics[width=\linewidth]{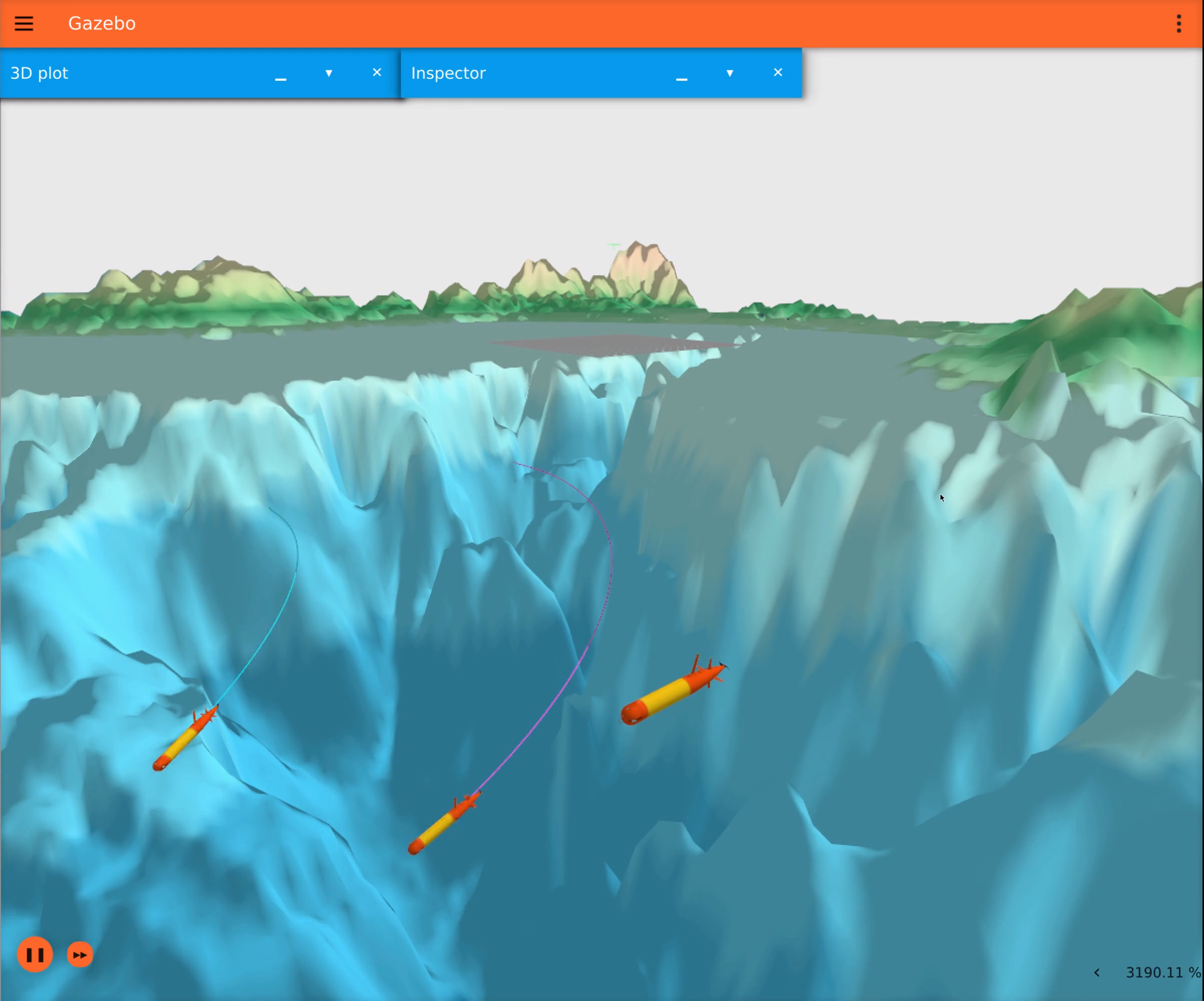}
      \caption{}
      \label{fig:multi_lrauv}
    \end{center}
  \end{subfigure}
  \begin{subfigure}{.52\linewidth}
    \begin{center}
      \includegraphics[width=\linewidth]{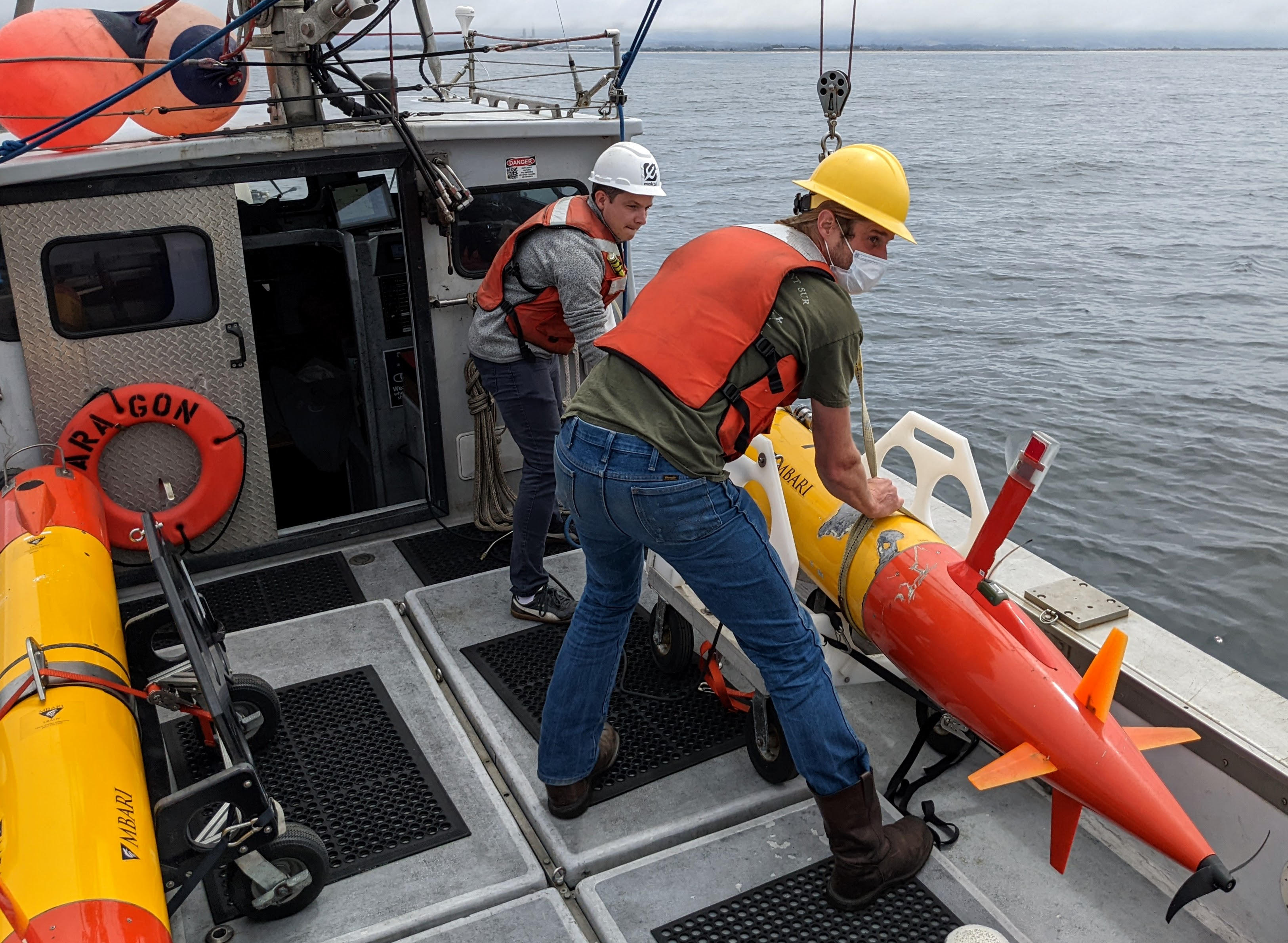}
      \caption{}
      \label{fig:deployment}
    \end{center}
  \end{subfigure} \hfill
  \vspace{-3mm}
  \caption{(a) Multi-vehicle simulation. (b) LRAUVs during field deployment from R/V \textit{Paragon} in Monterey Bay.} 
  \vspace{-5mm}
\end{figure}

While real-world validation is still necessary for mission development, the ability to rapidly simulate diverse scenarios allows practitioners to focus time in the field on fine-tuning missions to account for elements that are imperfectly modeled in simulation, such as subtle hydrodynamic behaviors and sensor and actuator characteristics, instead of discovering software bugs in mission logic or control flow in the field. 

We validated the simulator through \ac{CI} tests and by developing, from simulation to successful field trials, a complex behavior to maintain observation by a multi-robot team. In the behavior, an \ac{AUV} uses acoustic localization and communication to precisely replace another \ac{AUV}, allowing the relieved vehicle to be recharged or re-dispatched: a monitoring technique relevant to ocean researchers.

Our contributions are as follows:
\begin{itemize}
    \item Multi-robot \ac{FTRT} marine simulation, with fastest \ac{RTF} we know of
    \item Continuous controller integration for hydrodynamics validation based on solid theoretical foundations
    \item Synchronization of physics simulation time step with iterations of black-box controller
    \item Simulation-accelerated mission development and validation demonstrated in the real world
    \item In-simulator visualization of large-scale dense data, including dynamically interpolated science data, and high-resolution real-world bathymetry
    \item Software contribution accepted as native built-in features in a general simulator (the new Gazebo \cite{newgazebo})
\end{itemize}

\section{Related Work}




As with most underwater simulators, 
\ac{LRAUV} Sim provides typical building blocks, such as hydrodynamics, acoustic sensors, and marine-specific actuators.
Different from existing simulators, \ac{LRAUV} Sim is unique in its
stress-tested \ac{FTRT} speedup, 
field-tested multi-vehicle mission simulation, 
meticulously verified synchronization of physics time steps with the real control loop,
continuous integration with the controller, 
and visualization tools to assist development.

In short, \ac{LRAUV} Sim is a mission-driven validation tool that has been rigorously tested against a real controller and mission-validated at sea.
Further, it is not only an underwater simulator, but also a demonstration of software modules that have been accepted as built-in features to a widely used open source general robotics simulator (the new Gazebo).

\begin{table}[b]
  \renewcommand{\arraystretch}{1.2}%
  \begin{tabularx}{\columnwidth}{@{} p{2.2cm} p{2.3cm} *{3}{X} @{}}
  \toprule
  \textbf{Underwater Extensions} &
    \textbf{Simulator Platforms} &
    \textbf{Multi-Agent} &
    \textbf{Acoustic Comms} &
    \textbf{Large Worlds}
    \\ \midrule
  LRAUV Sim & (new) Gazebo \cite{newgazebo} & \cmark & \cmark & \cmark \\
  HoloOcean \cite{potokar2022_unreal} & Unreal Engine \cite{unreal} & \cmark & \cmark & \xmark \\
  Aqua \cite{manderson2018_unreal} & Unreal Engine \cite{unreal} & \xmark & \xmark & \xmark \\
  Stonefish \cite{cieslak2019} & Bullet Physics \cite{coumans2015} & \xmark & \xmark & \xmark \\
  URSim \cite{katara2019_unity} & Unity \cite{unity} & \xmark & \xmark & \xmark \\
  DAVE \cite{Zhang2022} & Gazebo-classic \cite{gzclassic} & \cmark & \xmark & \cmark \\
  UUV Sim \cite{manhaes2016uuv} & Gazebo-classic \cite{gzclassic} & \cmark & \xmark & \xmark \\
  \verb|ds_sim| \cite{vaughn_dssim} & Gazebo-classic \cite{gzclassic} & \cmark & \cmark & \cmark \\
  UWSim-NET \cite{Centelles2019} & Gazebo-classic \cite{gzclassic} & \cmark & \cmark & \xmark \\
  Freefloating \cite{kermorgant2014} & Gazebo-classic \cite{gzclassic} & \xmark & \xmark & \xmark \\
  Suzuki et al. \cite{suzuki2020} & Choreonoid \cite{choreonoid} & \xmark & \xmark & \xmark \\
  MARS \cite{tosik2016} & Bullet Physics \cite{coumans2015} & \cmark & \cmark & \xmark \\
  UW MORSE \cite{henriksen2016_uwmorse} & MORSE \cite{morse} & \xmark & \cmark & \xmark \\
  Rock \cite{watanabe2015_rock} & Gazebo-classic \cite{gzclassic} & \xmark & \xmark & \xmark \\
  Melman et al. \cite{melman2015} & & \xmark & \cmark & \xmark \\
  uSimMarine \cite{benjamin2018_moos_sim} & MOOS-IvP \cite{moosivp} & \xmark & \xmark & \xmark \\
  Sehgal et al. \cite{sehgal2010_modeling} & USARSim \cite{usarsim2007} & \cmark & \cmark & \xmark \\
  Song et al. \cite{song2003} & POSIX & \cmark & \cmark & \xmark \\
  DVECS \cite{choi2000} & Yuh et al. \cite{yuh1992} & \cmark & \xmark & \xmark \\
  \end{tabularx}
  \caption{Comparison of underwater simulation packages}
  \label{tbl:sims}
\end{table}

\ac{FTRT} and multi-vehicle simulation are not always a high priority, because underwater simulation is often used for brief missions such as manipulation and inspection.
\ac{FTRT} is necessary for complex missions with long-term autonomy, which require validating event sequences that lead to failures.
We measure \ac{FTRT} by the \ac{RTF}, which is 1 at real time, $<1$ for slower, and $>1$ for faster.
Multi-vehicle simulation is necessary for tasks such as collaborative seafloor mapping \cite{manhaes2016uuv} and environmental monitoring.
To validate \ac{LRAUV} Sim for multi-vehicle use cases, we tested for stable physics performance at \ac{RTF} 100 for a single vehicle, and we measured the \ac{RTF} while increasing the number of vehicles.

Table \ref{tbl:sims} compares several modern underwater simulations on features relevant to our contributions.
UWSim \cite{prats2012_uwsim} was among the first to extend the popular Gazebo-classic \cite{koenig2004design, gzclassic} simulator with underwater capabilities, by implementing hydrodynamics 
for manipulation with intervention \acp{AUV}.
UWSim was extended and studied in \cite{kermorgant2014, Centelles2019, fernandez2015, gwon2017}.

In a newer generation, \ac{UUV} Simulator \cite{manhaes2016uuv} was developed beyond the basic hydrodynamics, sensors, and actuators for Gazebo-classic. Most related to our work, it simulated multi-vehicle seabed mapping, chemical plumes, and a particle concentration sensor.
DAVE \cite{Zhang2022} extended \ac{UUV} Simulator by adding a CUDA-enabled physics-based multi-beam sonar \cite{choi2021multibeam}, water tracking for the \ac{DVL}, dynamic tile loading for large-scale worlds, dual-arm manipulation, and other features.
AUG Sim \cite{choi2022_epicdaug} simulated an \ac{FTRT} glider. However, it did not demonstrate multiple vehicles.

\ac{LRAUV} Sim is based on the new Gazebo \cite{newgazebo}, which is entirely rewritten and completely different from Gazebo-classic \cite{gzclassic}.
Similar to \cite{manhaes2016uuv} and \cite{Zhang2022}, state-of-the-art holistic underwater simulators, we implemented hydrodynamics using Fossen's equations \cite{Fossen1994}.
However, though \ac{FTRT} was possible in Gazebo-classic, these works did not rigorously measure the extent of speedup possible, especially for multiple vehicles. To do so, we vary the physics timestep while using an automated test suite to measure navigation trajectory errors. For instance, we numerically verify that the modeled vehicle can execute motions (both open-loop and trajectories commanded by a real controller) within deltas from a reference trajectory as the real-time factor increases. This test suite is explained further in Section \ref{sec:continuous_integration}.
\ac{UUV} Simulator and DAVE did not have continuous integration test suites validated with a real controller.

Further, their worlds and science data were not demonstrated at as large a scale as \ac{LRAUV} Sim ($150$ miles).
Large worlds are inherent to marine environments and pose a simulation performance challenge.
Note that the home-brewed dynamic tiling for large worlds, which DAVE \cite{Zhang2022} adapted from \verb|ds_sim| \cite{vaughn_dssim}, is a built-in feature (Levels) in the new Gazebo, which \ac{LRAUV} Sim demonstrated through high-definition bathymetry.


HoloOcean \cite{potokar2022_unreal} was based on Unreal Engine \cite{unreal}, a 3D graphics game engine. 
Targeting the visuals for SLAM applications, it did not emphasize accurate hydrodynamics.
Gazebo is known to have more accurate physics than game engines, as it prioritizes high-fidelity physics engines for robotics.
We modeled all the actuators of a robot deployed at sea more than 300 days a year. Thus, hydrodynamics' impact on vehicle behavior was important. Using an automated test suite, we rigorously tested the interactions between actuators and hydrodynamics.

Conventionally, simulation data are either plotted offline from log files, or visualized dynamically in a separate tool.
We developed native GUI components to overlay point clouds directly in the simulation, making visualization and debugging more efficient. 

We emphasize that though our test bed was a specific vehicle, our software contributions have been accepted as native features in the open source Gazebo simulator and are available as individual modules to general robotics.
Though we synchronized the physics time step to a specific controller, we demonstrated that low-level coupling enables simulation to be used in a high-fidelity manner to quickly validate long-term mission success using real-world controllers.

\section{Dynamics Formulation}


Basic dynamics and common actuators are modeled using standard methods. We summarize here for completeness.

\subsection{General Hydrodynamics}
\label{sec:hydro_theory}

We model basic hydrodynamics using Fossen's equations \cite{Fossen1994}:
\begin{equation}
  M\ddot{x} + D(\dot{x})\dot{x} + C(\dot{x})\dot{x} = F \label{eq:1}
\end{equation}
where $M$ is the added mass matrix, $D(\dot{x})$ is the damping matrix, $C(\dot{x})$ is the Coriolis matrix, and $\dot{x}$ is the velocity six-vector. Ocean current is approximated by adding a term $v_c$ to the velocity:
\begin{equation}
  M\ddot{x} + D(\dot{x}-v_c)(\dot{x}-v_c) + C(\dot{x}-v_c)(\dot{x}-v_c) = F\label{eq:2}
\end{equation}
In our case, $C(\dot{x})$ is small, so $D(\dot{x})$ is the main contribution. We only simulate the diagonal terms:
\begin{equation}
  \begin{aligned}
    D(\dot{x}) = &-\text{diag}\{X_u, Y_x, Z_w, K_p, M_q, N_r\}
    \\ &- \text{diag}\{ X_{u|u|}|u|, Y_{v|v|}|v|, Z_{w|w|}|w|,
    \\ & K_{p|p|}|p|, M_{q|q|}|q|, N_{r|r|}|r| \}
  \end{aligned} \label{eq:3}
\end{equation}

\subsection{Thruster}
We model the thruster force $F$ after \cite{Fossen1994} as well:
\begin{equation}
  F = \rho D^4 K_T(J_0) n |n|\label{eq:thrust}
\end{equation}
where $D$ is the fin's diameter. $n$ is rotations per second, and $K_T(J_0)$ is the propeller's coefficient, empirically determined.

\subsection{Lift on Fins}
Fins are common on gliders. When the fins are angled, a lifting force is generated, which causes the vehicle to rotate. These are modeled using the classic lift equation:
\begin{equation}
  F = \frac{1}{2} C_l \rho v^2 A \label{eq:lift}
\end{equation}
where $C_l$ is the lift coefficient, $v$ is the velocity, and $A$ is the area.  We model the value of $C_l$ as a piece-wise linear function. $C_l$ varies linearly with the angle of attack until the stall angle is reached.



\section{Simulator Features}

\begin{figure*}[tb]
  \begin{subfigure}{.32\linewidth}
    \begin{center}
      \includegraphics[height=4cm]{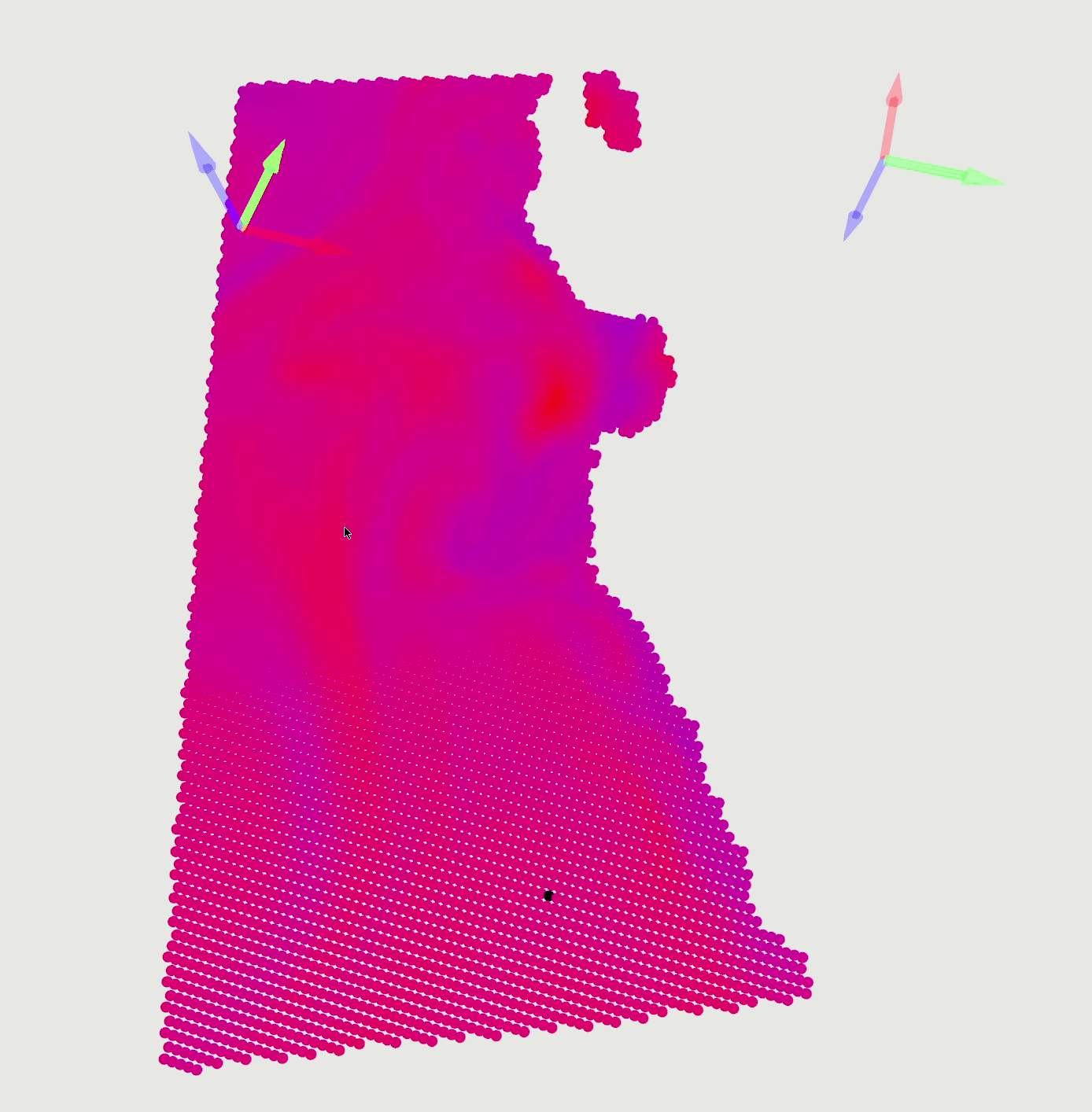}
      \caption{}
      \label{fig:sci_top}
    \end{center}
  \end{subfigure}
  \begin{subfigure}{.32\linewidth}
    \begin{center}
      \includegraphics[height=4cm, width=4cm]{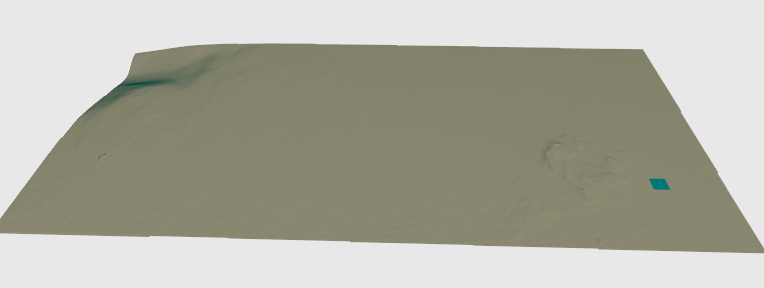}
      \caption{}
      \label{fig:bathy}
    \end{center}
  \end{subfigure}
  \begin{subfigure}{.32\linewidth}
    \begin{center}
      \includegraphics[height=4cm]{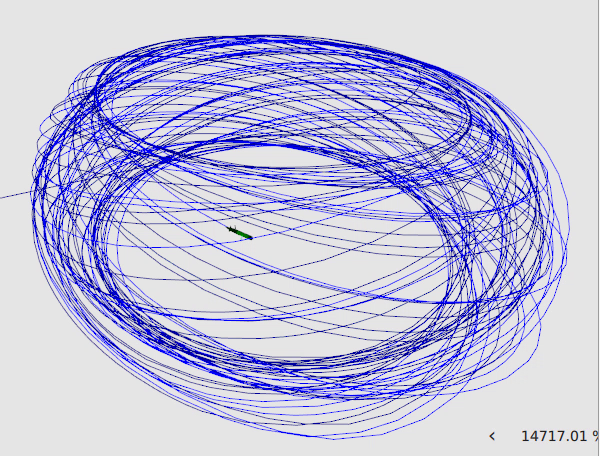}
      \caption{}
      \label{fig:yoyo}
    \end{center}
  \end{subfigure}
  \vspace{-4mm}
  \caption{Features relevant to robot missions for oceanography, navigation, or controls. (a) Visualization of large-scale dense data on sea temperature in coastal California, from San Francisco Bay at the top, Monterey Bay in top middle, and near San Simeon at the bottom, spanning $\sim$150 miles. Resolution was variable, finest being 5 \si{\meter} in z depth. Red: high temperature; blue: low. (b) Detailed bathymetry, with 1 \si{\meter} resolution, of Portuguese Ledge in Monterey Bay. (c) 3D plot of robot trajectory in the Circling Yo-Yo mission.}
  \vspace{-5mm}
\end{figure*}

Various aspects relevant to ocean science mission development are modeled. We use the DART physics engine \cite{Lee2018}, but others can be used. Data such as temperature and salinity can be queried and visualized, high-resolution bathymetry can be dynamically loaded, and inter-vehicle communication is modeled. Code from black-box controllers is integrated with the simulator through inter-process communication. These features provide a framework for multi-robot mission development.

\subsection{Volumetric Data}
Historic data, such as ocean current, salinity, temperature, and chlorophyll, that vary spatially with arbitrary resolution can be loaded and queried. For roboticists, changes in current, temperature, and density affect vehicle dynamics and communications. For biologists, they affect ocean ecology.

\subsubsection{Environmental Data Interpolation}
Environmental data like current, temperature, or chlorophyll are often generated by predictive models at a spatial scale of kilometers or degrees, while the spatial scale relevant to simulating \ac{AUV} dynamics and sensors is local. To achieve local observations, we store environmental data in a sparse grid and trilinearly interpolate among the nearest points on the grid.

The main challenge is to efficiently query grid data. The grid is not uniformly spaced, due to conversion errors between spherical coordinates and Mercator projections. Often, the depth axis has logarithmic spacing between data points. This makes modeling it as a simple 3D array precarious.

To handle these inconsistencies in resolution, we use a special data structure. The axes are indexed as red-black trees in memory and queried using binary search. Binary search on this axis returns the closest points in $O(\log n)$ time, where $n$ is the number of points on the axis. If the value is found, it is returned; otherwise, it returns the closest 2, 4, or 8 points, depending on whether the query lies on a plane or in between points in the grid. We then utilize linear, bilinear or trilinear interpolation to estimate the value of the current point.

Linear interpolation is further used between values that change in time, to estimate values between time steps. We extend this to time-varying 3D data by loading and interpolating between the previous and next data time step. Searching a 1 GB data file takes on the order of 10--100 \si{\nano\second} per query on a laptop with a AMD Ryzen 5900HX CPU.

\subsection{GUI and Visualization Tools}

\subsubsection{Point Cloud Visualization of Environmental Data}
Data visualization aids in development of oceanographic sampling missions. Fig. \ref{fig:sci_top} 
shows temperature distribution in Monterey Bay from historical data. We provide real data samples for chlorophyll, temperature, salinity, and ocean current. Multiple types of data can be overlaid simultaneously in configurable colors. The implementation allows the user to easily extend to more data types.

\subsubsection{3D Trajectory Plot}
We perform 3D visualization of the robot's trajectory natively in the simulator, useful for debugging control and navigation algorithms. Fig. \ref{fig:yoyo} shows the robot's trajectory in a Circling Yo-Yo mission.

\subsection{High-Resolution Bathymetry}
Real-world high-resolution (1 \si{\meter}) bathymetry (Fig. \ref{fig:bathy}) collected by vehicles can be loaded while maintaining simulation performance. We split the bathymetry into tiles and only load tiles directly below the vehicle. This allows for modeling large-scale terrain while maintaining low latency, useful for developing algorithms that use bathymetry to navigate or to determine sampling locations.

\subsection{Acoustic Communication}
In a modular architecture, we use a delay-based protocol with a range limited drop-off model. A simple model meets our needs, but the modular architecture allows for more complex acoustic models. The acoustic communication model also enables an acoustic localization plugin, allowing robots to query the relative position of an acoustic beacon.

\subsection{Integration with Controller}
The vehicle control loop is synchronized with the simulated physics loop by reading a simulation timestamp from the simulator state. As the physics time step, or simulation speed, goes faster in real-world time, the control loop issues commands faster.

To achieve \ac{FTRT}, large time steps are used, e.g., 20 \si{\milli\second} for navigation as opposed to the Gazebo default of 1 \si{\milli\second}. Lower physics fidelity that accompanies larger time steps is compensated by a closed-loop controller \cite{willsky1997signals} and verified to be within acceptable delta bounds using automated tests.

\begin{figure}[thp]
    \center
    \includegraphics[width=0.8\linewidth]{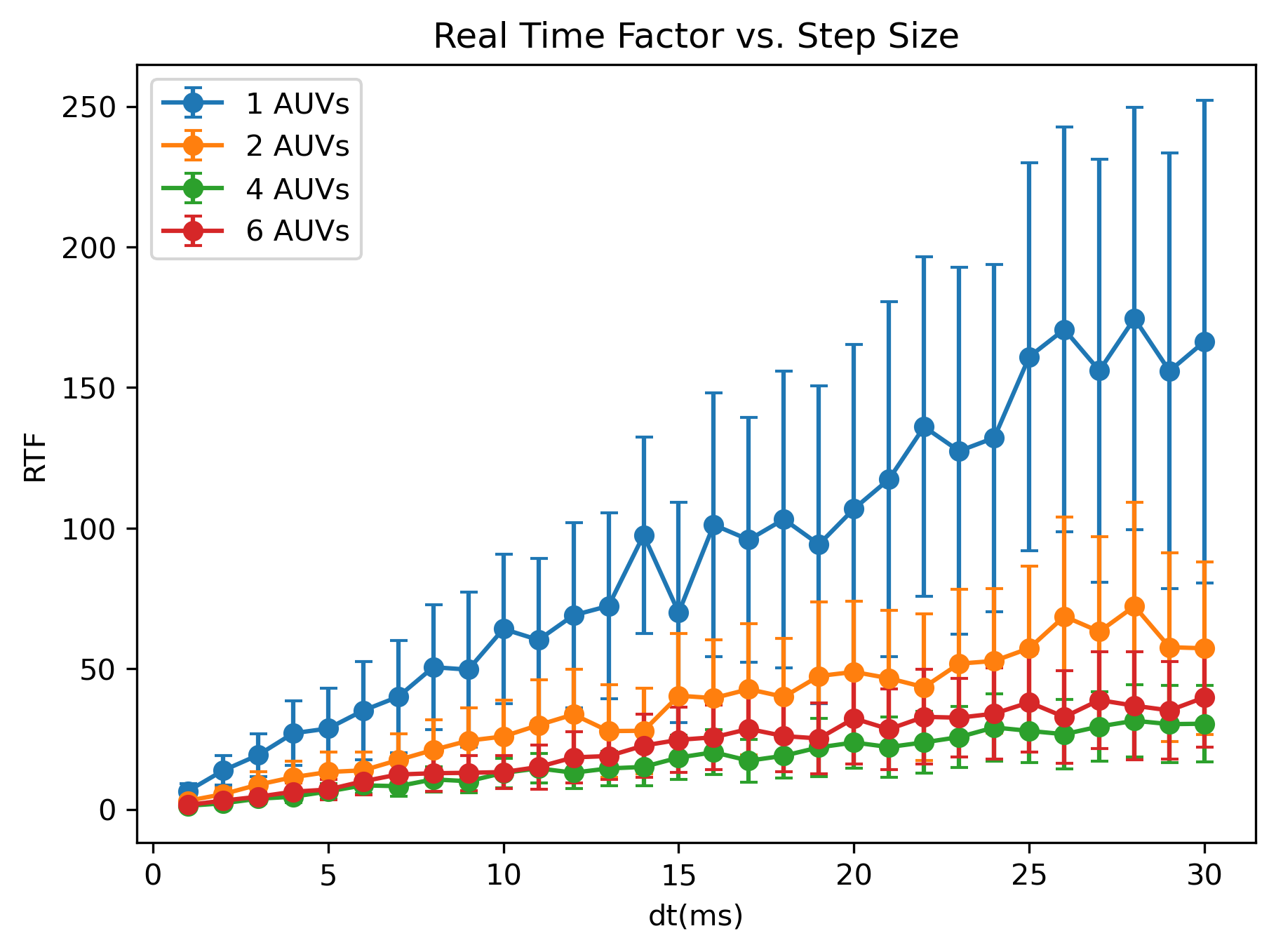}
    \caption{\ac{RTF} increased with physics step size and was inversely proportional to the number of vehicles during a Circling Yo-Yo mission.}
    \label{fig:rtf}
    \vspace{-5mm}
\end{figure}

Fig. \ref{fig:rtf} shows the \ac{RTF} performance while increasing the number of vehicles, up to a physics time step of 30 ms. The \ac{RTF} is inversely proportional to the number of vehicles. On a single vehicle, we achieved \ac{RTF} $>100$, or 100 times \ac{FTRT}, at a time step of 30 \si{\milli\second}. The robot was still able to perform the Circling Yo-Yo mission shown in Fig. \ref{fig:yoyo}, indicating sufficient physics fidelity for mission-level verification.



\section{Validation through Continuous Integration}
\label{sec:continuous_integration}

In implementing the hydrodynamics, lift-drag, buoyancy, and other fundamental factors that affect the physics of underwater vehicles, it was necessary to validate each actuator and its physics effect individually, before combining all of them. Otherwise, it was impossible to track down which part of the theoretical formulation was incorrectly implemented.

This was carried out by \ac{CI} testing. Each time the simulator is changed, such as through updating the source code, an automated test suite is rerun to verify correctness, to make sure new additions do not break existing behaviors. The suite consists of several types of tests:
\begin{itemize}
    \item Unit tests of basic physics to verify individual forces (e.g., buoyancy on a geometric primitive)
    \item Unit tests of physics with simple actuator commands (e.g., a translation on the battery prismatic joint), for vehicle model correctness
    \item Integration tests of the simulated vehicle with real-world controller on individual actuators (e.g., holding a constant value on the rudder to go in a circle), to verify synchronized control loop time steps
    \item Integration tests at the mission level (e.g., Circling Yo-Yo)
\end{itemize}

\begin{figure*}[htbp]
    \center
    \includegraphics[width=0.9\linewidth]{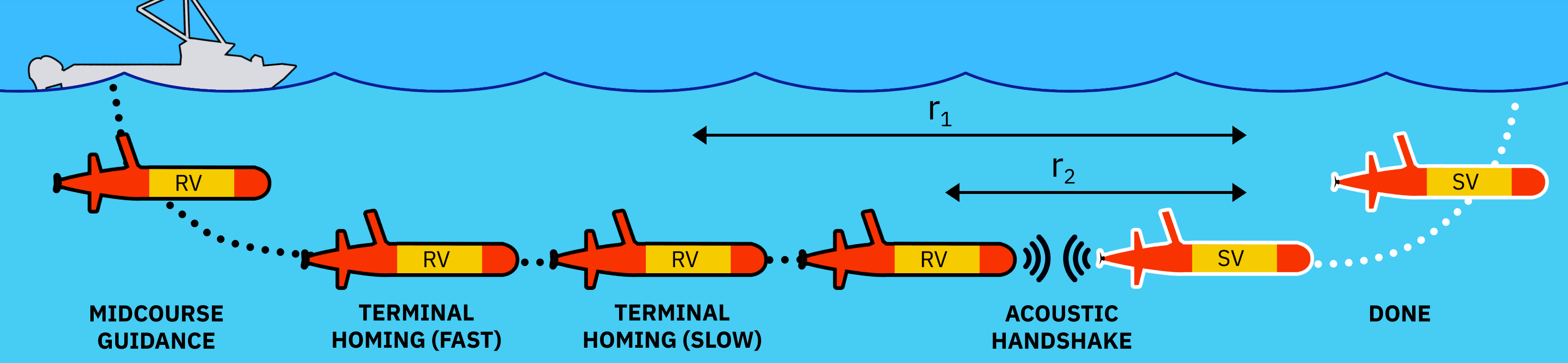}
    \caption{Hot-bunking mission phases. Once deployed, the relief vehicle (RV) begins a three-stage homing sequence toward the drifting sampling vehicle (SV): after achieving a GPS waypoint during \textit{Midcourse Guidance}, the RV switches to an acoustic \textit{Terminal Homing} stage with fast speed before slowing down once range is reduced to $r_1$. Once a range of $r_2$ is attained, a two-way \textit{Acoustic Handshake} occurs where the RV commands the SV to surface and the SV acknowledges the command and surfaces. In the \textit{Done} phase, the RV begins sampling.}
    \label{fig:phases}
    \vspace{-6mm}
\end{figure*}

\subsubsection{Tests of the Vehicle Model without the Controller}
These tests explore a minimum set of invariants that a simulated underwater vehicle should satisfy. We test these invariants and compare them to real world constants. Though the test tolerances are tuned for one vehicle, the components tested, such as hydrostatics and hydrodynamics, and the actuators tested, such as propellers and fins, are common among underwater gliders.


\paragraph{Hydrostatics} 
We ensure the vehicle model is neutrally buoyant and has form stability (i.e., it tends to right itself). For neutral buoyancy, we test that the vehicle does not move when starting at rest and external forces are absent.

For stability, we enable buoyancy and gravity without hydrodynamic drag and start the vehicle at some offset angle $\theta_0$. The off-center buoyancy results in a restoring moment given by
\begin{equation}
  \tau(\theta) = \rho g V C_b \sin\theta \label{eq:restoring_moment}
\end{equation}
where $V$ is volume, $C_b$ is the vertical displacement of the center of buoyancy from the center of mass, and $g$ is the gravitational acceleration. We check that this results in an oscillation between $-\theta_0$ and $\theta_0$, which is expected by conservation of energy. 


\paragraph{Hydrodynamics}
To verify hydrodynamics, we check that the vehicle velocity is realistically damped. Since the hydrodynamics equations (Section \ref{sec:hydro_theory}) consist of one acceleration- and two velocity-dependent terms, it is necessary to test the terms separately. 
\begin{equation}
  \underbrace{M\ddot{x}}_{\text{Acceleration-dependent term}} + \underbrace{D(\dot{x})\dot{x} + C(\dot{x})\dot{x}}_{\text{Velocity-dependent terms}}
\end{equation}
For the acceleration-dependent term, we monitor response to oscillations. For a naturally stable vehicle, the acceleration term should damp oscillations. We check this by introducing a position-dependent force to induce oscillations. This should result in a scenario similar to simple harmonic oscillation; an unstable oscillation would fail the automated test.

We check the velocity-dependent terms by simulating the vehicle at maximum thrust. Given a constant force, the equations of motion should converge \cite{Fossen1994} to a certain terminal velocity. Experimentally, we know this value to be 1 m/s. We verify that a similar terminal velocity is reached.

\paragraph{Vertical Fins (Rudder)}
Fins introduce forces that cause the vehicle to pitch or yaw. We test the horizontal and vertical fins separately. For the vertical fins, we perform a unit test by moving the vehicle in a circular trajectory. The force exerted by the fins is given by Eqn. \eqref{eq:lift}, which we equate to the expression for circular motion:
\begin{equation}
     \frac{1}{2} C_l \rho v^2 A = \frac{mv^2}{r} \label{eq:unsimplified_circ}
\end{equation}
This further simplifies to:
\begin{equation}
    r = \frac{2m}{C_l \rho A} \label{eq:assert_circ}
\end{equation}
While we could check the radius of curvature, floating point errors may accumulate. Hence, we check that the yaw rate during circular motion is a constant value of $\frac{2\pi r}{v}$.

\paragraph{Horizontal Fins (Elevator)}
When pitched, they exert a force that pitches the vehicle. The force is counteracted by the restoring moment in Eqn. \eqref{eq:restoring_moment}. We model the torque $\tau(\theta)$ and the fin's displacement $d$ from the center of buoyancy:
\begin{equation}
    \tau(\theta) = \rho g V \sin \theta - \frac{1}{2}C_l\rho v^2 A d \label{eq:fin_moments}
\end{equation}
Solving Eqn. \eqref{eq:fin_moments} for $\tau(\theta) = 0$ gives the min/max pitch the vehicle can have when traveling at speed $v$. At 1 m/s, the vehicle pitches a maximum of $\pm20^{\circ}$. This is another invariant used to validate the model.

\paragraph{Battery Mass Shifter}
A vehicle-specific actuator, the mass shifter, trims the center of mass by moving the battery to pitch the vehicle \cite{Hobson2012}. To test it, we translate the mass to a fixed location. This introduces a torque, which is countered by the buoyancy force. This is similar to horizontal fins (Eqn. \eqref{eq:fin_moments}), except here $d$ is variable, and $m_{\text{shifter}}$ represents the mass shifter's weight:
\begin{equation}
    \tau(\theta) = \rho g V C_b \sin \theta - m_{\text{shifter}} \, g \, d \, \cos(\theta) \label{eq:mass_shifter}
\end{equation}
Solving for $\tau(\theta) = 0$ gives the equilibrium pitch angle for a given $d$. This value is verified against an empirical reference.

\subsubsection{Integration Tests with the Controller}
We developed integration tests with the real vehicle's controller, testing that each component works correctly. At the unit integration level, for instance, we check that shifting the mass causes the vehicle to pitch. At the mission integration level, for instance, the vehicle performs a Circling Yo-Yo mission. This involves controlling the vehicle to circle downward toward the sea floor and then resurface multiple times. We check that the vehicle's direction of travel and yaw rate are sufficiently close to the reference trajectory.




\section{Experimental Results}
We validated the simulator by developing and testing a real multi-robot mission called ``acoustic hot-bunking," where an \ac{AUV} uses acoustic localization to replace another \ac{AUV} at its precise location. Hot-bunking extends the duration that a specific water mass can be monitored, which is needed to study drifting microbial populations \cite{zhang2021system}, phytoplankton patch formation \cite{jaffe2017swarm}, and the impact of floating sediment on cellular life \cite{ryan2014boundary}.

This mission is challenging to develop because failures in either robot could occur at each of the five mission phases shown in Fig. \ref{fig:phases}, requiring simulated tests to ensure that the two-robot system can recover from errors in each phase. The recovery behaviors would be impractical to test in a real-time simulated deployment of over two hours. We used the simulator's capability for multi-robot simulation, acoustic communications, and integration with the real-world controller to test these scenarios in \ac{FTRT}.

\subsection{Mission Definition}


In the mission, the \ac{SV} is replaced by the \ac{RV} using the sequence of phases shown in Fig. \ref{fig:phases}. To ensure that the \ac{SV} is replaced in its exact location despite odometry drift in a GPS-denied environment \cite{woodman2007introduction}, we perform closed-loop homing using relative localization from a \ac{DAT}. Homing is done using a pure pursuit controller, such that
\begin{equation}
  \theta_{des} \Leftarrow \theta_{SV \leftarrow RV, \, meas}
\end{equation}
where $\theta_{des}$ is the commanded yaw and $\theta_{SV \leftarrow RV, \, meas}$ is the azimuth from the RV to the SV, in a global frame, measured by the \ac{RV}'s acoustic transponder.




\subsection{Mission Validation}


Validation involved testing the nominal event sequence, where each mission phase worked as intended, and the off-nominal event sequences where failures must be accommodated. To verify the nominal event sequence, we developed unit tests for each phase to ensure successful operation.

However, in field deployments, vehicle capability could be compromised by environmental factors (e.g., seaweed entangling a propeller) or software factors (e.g., a bug disabling one vehicle’s acoustic modem). Consequently, we also simulated component failures and verified that appropriate timeout behaviors were triggered if components failed in any of the phases. \ac{FTRT} simulation allowed each test to complete in minutes rather than over two hours in real time.

\subsection{Results}

The hot-bunking mission was run on two vehicles: \textit{Brizo}, the \ac{SV}, and \textit{Tethys}, the \ac{RV} (Fig. \ref{fig:deployment}). Both were equipped with Teledyne Benthos \acp{DAT} which provided relative acoustic localization and underwater communications. The vehicles were deployed from the R/V \textit{Paragon} for a two-day saltwater deployment above Portuguese Ledge in Monterey Bay, during which three hot-bunking trials were executed via Iridium satellite connection.

The field trials enabled high-level parameters to be tuned from field data that was not possible in simulation. The first trial inspired an adjustment of the slow terminal homing speed from $0.5$ m/s to $0.8$ m/s, after it was observed that a low speed was insufficient to maintain control authority in the real system.

In the second trial, the two-way acoustic handshake (Fig. \ref{fig:phases}) was only partially successful: the \ac{RV} did not receive an acknowledgement signal from the \ac{SV} because the acoustic bandwidth was overshadowed by localization requests from the \ac{DAT}. In the next trial, the frequency of acoustic localization was modified to prevent this issue, which was not present in simulation. The third trial successfully demonstrated all phases of the hot-bunking sequence.

\subsubsection{Acoustic Homing}
\begin{figure}[!t]
    \center
    \includegraphics[width=\columnwidth]{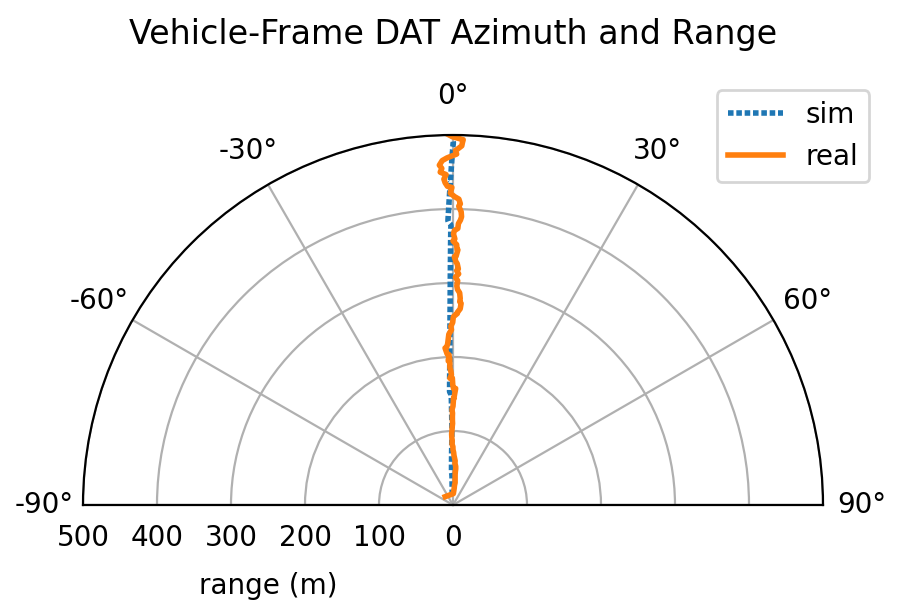}
    \caption{Polar trace of the $xy$-plane \ac{DAT} vector in the relief vehicle's frame, in simulation and reality. The range to the sampling vehicle decreased until the 20 \si{\meter} success criteria. }
    \label{fig:polar}
  \vspace{-5mm}
\end{figure}

Fig. \ref{fig:polar} shows the trajectory of the \ac{DAT} fix in the \ac{RV} frame during the third trial, advancing from a range of $500$ \si{\meter} to within the success radius of $20$ \si{\meter}, while keeping the \ac{SV} approximately straight ahead of the \ac{RV} in both simulated and real trials. The decreasing trend in range in both simulated and real trajectories indicates the successful operation of the control law.

The real trajectory (orange) displays more oscillation than the simulated trajectory (dotted blue). The oscillation in \ac{DAT} azimuth can be attributed to hydrodynamic forces in the ocean inducing low-amplitude yawing motions to a greater extent than in simulation. 

\subsubsection{Acoustic Handshake}
In the third trial, the acoustic handshake was successfully carried out: the \ac{RV} sent a ``relieved-of-duty" transmission; the \ac{SV} received and acknowledged it; the \ac{RV} registered the acknowledgement and began scientific sampling, and the \ac{SV} ascended to the surface. The acoustic communications behaviors developed in simulation were successful on the real system.

\section{Discussion}
\ac{LRAUV} Sim enabled fast iteration during mission development, to quickly bring a novel and impactful \ac{AUV} behavior to fruition. While two robots constitute the minimum of a multi-robot system, the benefits of \ac{FTRT} simulation for our field trials should scale to greatly benefit larger fleets. The simulations preceded the field trials, so that development time in the field was spent where it should be: tuning mission parameters based on field data rather than troubleshooting software components and mission control flow.

By supporting development with simulations of many event sequences, we caught multiple bugs in control flow, including ones infeasible to identify without the broad test coverage enabled by \ac{FTRT} simulation -- a crucial benefit to larger fleets with more failure cases.

\section{Acknowledgements}
We are grateful for support from the David and Lucile Packard Foundation. This work would not have been possible without
the team at Open Robotics and Ekumen Labs.

\addtolength{\textheight}{-2cm}   

\bibliographystyle{IEEEtran}
\bibliography{references}

\end{document}